\documentclass[conference]{IEEEtran}
\IEEEoverridecommandlockouts
\usepackage[caption=false,font=normalsize,labelfont=sf,textfont=sf]{subfig}
\usepackage{graphicx}
\usepackage{standalone}
\usepackage{tikz}
\usetikzlibrary{shapes,backgrounds,shapes.geometric, arrows, arrows.meta, bending, positioning}
\usepackage{circuitikz}
\tikzset{font={\fontsize{9pt}{9}\selectfont}}
\usepackage{pgfplots}
\usepgfplotslibrary{fillbetween,colormaps,groupplots}
\pgfplotsset{compat=newest}
\usepackage{xcolor}
\usepackage{pgfplotstable}
\usepackage{pgffor}
\usepackage{ifthen} 
\usepackage{multirow}
\usepackage{datatool}

\usepackage{tabularx}
\usepackage{makecell}
\usepackage{booktabs}
\newcolumntype{Y}{>{\centering\arraybackslash}X}
\newcolumntype{s}{>{\hsize=0.7\hsize}Y}

\usepackage{siunitx}
\usepackage{amsmath}
\usepackage{amsfonts}
\usepackage{dsfont}
\usepackage{scalerel}
\usepackage{amssymb}
\usepackage{bm}

\usepackage{algorithm}
\usepackage{algorithmicx}

\usepackage{cite}
\usepackage{jabbrv}

\usepackage[utf8]{inputenc}
\usepackage{enumitem}

\date{\today}

\usepackage[
	pdfauthor={Julian Schumann},
	pdfcreator={pdfLatex},
	pdfsubject={First project for PhD},
    colorlinks={true},
    linkcolor={blue},
    citecolor={blue},
    urlcolor={red},
    hyperindex={true}
]{hyperref}
\makeatletter
\hypersetup{pdftitle=\@title}
\makeatother
\usepackage[margin=0.75in, top=1in]{geometry}
\usepackage{textcomp}
\def\BibTeX{{\rm B\kern-.05em{\sc i\kern-.025em b}\kern-.08em
    T\kern-.1667em\lower.7ex\hbox{E}\kern-.125emX}}

\title{Evaluating randomized smoothing as a defense against adversarial attacks in trajectory prediction}

\author{
    \IEEEauthorblockN{Julian F. Schumann\IEEEauthorrefmark{1}, Eduardo Figueiredo\IEEEauthorrefmark{2}, Frederik Baymler Mathiesen\IEEEauthorrefmark{2},\\ Luca Laurenti\IEEEauthorrefmark{2}, Jens Kober\IEEEauthorrefmark{1}, Arkady Zgonnikov\IEEEauthorrefmark{1}\\}
    \IEEEauthorblockA{\IEEEauthorrefmark{1}\textit{Department of Cognitive Robotics}, \textit{TU Delft}, Delft, Netherlands}
    \IEEEauthorblockA{\IEEEauthorrefmark{2}\textit{Delft Center for Systems and Control}, \textit{TU Delft}, Delft, Netherlands\\
    \{j.f.schumannn, e.figueiredo, f.b.mathiesen, l.laurenti, j.kober, a.zgonnikov\}@tudelft.nl}
    
    \thanks{The source code used in this work is available online at \href{https://github.com/julianschumann/General-Framework-Smoothing}{https://github.com/julianschumann/General-Framework-Smoothing}.}
}

\begin{document}
\twocolumn
\maketitle

\begin{abstract}
    Accurate and robust trajectory prediction is essential for safe and efficient autonomous driving, yet recent work has shown that even state-of-the-art prediction models are highly vulnerable to inputs being mildly perturbed by adversarial attacks. Although model vulnerabilities to such attacks have been studied, work on effective countermeasures remains limited. In this work, we develop and evaluate a new defense mechanism for trajectory prediction models based on randomized smoothing -- an approach previously applied successfully in other domains. We evaluate its ability to improve model robustness through a series of experiments that test different strategies of randomized smoothing. We show that our approach can consistently improve prediction robustness of multiple base trajectory prediction models in various datasets without compromising accuracy in non-adversarial settings.
    Our results demonstrate that randomized smoothing offers a simple and computationally inexpensive technique for mitigating adversarial attacks in trajectory prediction.
\end{abstract}
\section{Introduction}
Autonomous driving has emerged as a key automotive innovation, attracting billions in investment~\cite{holland-letz_mobilitys_2021}, driven by its potential to reduce traffic injuries and fatalities~\cite{EU2019future} as well as to generate substantial economic value~\cite{duboz2025scenarios}.
However, for the viability of autonomous vehicles, being safer than human drivers alone is insufficient; they must also drive efficiently and interact smoothly with human road users. Early autonomous vehicles were often overly cautious, which paradoxically created risks, such as being rear-ended~\cite{milford_self-driving_2020, sinha_crash_2021}. This arises from the uncertainty about other road users’ behavior, limiting safe control options. Accurate prediction models can reduce this uncertainty, enabling more efficient and assertive driving, anticipating conflicts, and yielding when appropriate~\cite{mozaffari_deep_2022, schumann_benchmarking_2023}, while also supporting realistic simulation of human behavior in virtual environments~\cite{montali_waymo_2023}.
Consequently, many distinct prediction models have been proposed in recent years, with a primary focus on deep learning methods. The employed architectures range from conditional variational autoencoders~\cite{salzmann_trajectron_2020, nayakanti_wayformer_2023} and normalizing flows~\cite{meszaros_trajflow_2024} to denoising diffusion models~\cite{ho2020denoising,bae2024singulartrajectory} and large language models~\cite{seff2023motionlm,xu2025trajectory}.

While these models generally achieve impressive prediction accuracy on well-defined benchmarks, their robustness -- although essential for public safety~\cite{hagenus2024survey} -- is explored far less thoroughly. One emerging method for evaluating robustness is based on adversarial trajectories, which are specifically generated to deceive the model's predictions as much as possible; a model’s ability to withstand such attacks is known as adversarial robustness~\cite{tocchetti2022ai}. Initial work on adversarial attack generation has shown promise in exposing system vulnerabilities~\cite{cao_advdo_2022, cao_robust_2023, zhang2022adversarial,fan2024adversarial}. However, these works typically compare the model predictions under adversarial perturbations against ground-truth future trajectories of benign agents, which might not be kinematically reachable from the perturbed past on which these predictions are conditioned. Recent work has introduced additional constraints ensuring kinematic reachability to make such comparisons more physically meaningful~\cite{schumann2025realistic}. However, even under those more constrained and meaningful attacks, though, most state-of-the-art models still show significant degradation in performance. Additionally, such adversarial attacks have been shown to be remarkably transferable between different models when trained on similar datasets~\cite{schumann2025realistic,schumann2025step}, further highlighting the importance of trajectory prediction robustness for real-world automated driving systems.

\begin{figure*}
    \centering
    \includegraphics{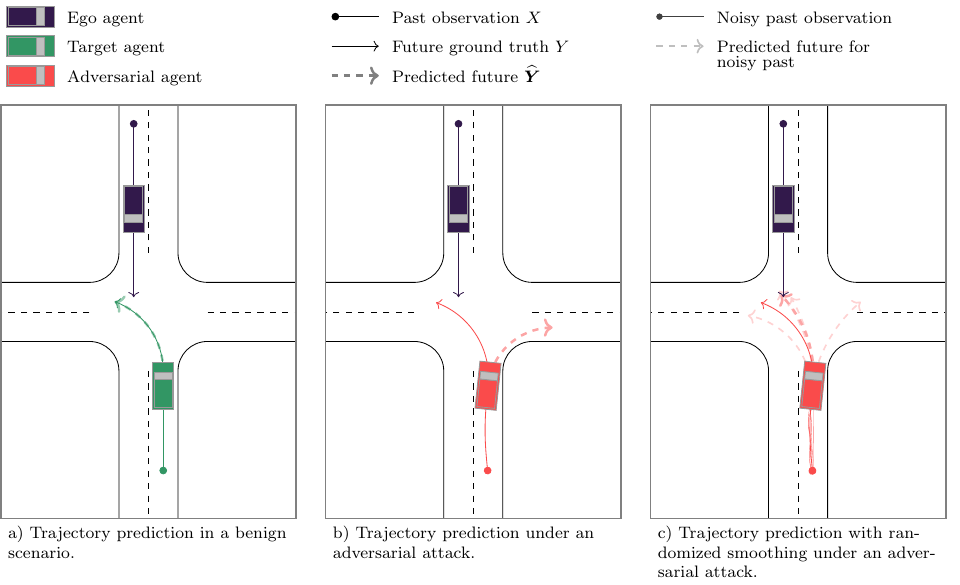}
    \caption{An overview of randomized smoothing applied to trajectory prediction. a) In a benign scenario, it can be expected that a well-trained trajectory prediction model $\mathds{P}$ can make predictions which align with the ground truth. b) However, if the target agent, for which predictions are made, uses an adversarial attack, the prediction quality can deteriorate even in state-of-the-art models~\cite{schumann2025realistic,schumann2025step}. c) Randomized smoothing can be used to overcome this issue. Instead of the actual past observations of the adversarial agent, the model is applied to multiple randomly perturbed versions of these observations. The final prediction is then built by averaging those separate model predictions.}
    \label{fig:overview}
\end{figure*}

While prior work has examined the design and evaluation of adversarial attacks~\cite{hagenus2024survey}, research on countering such attacks in trajectory prediction models is very limited. Recently, Fan \emph{et al.}  proposed a method for detecting adversarial trajectories~\cite{fan2024novel} which could enable an autonomous vehicle to avoid being misled by an adversarial agent. However, beyond this work, systematic approaches to defending trajectory prediction models against adversarial attacks remain largely unexplored. At the same time, outside trajectory prediction, \emph{randomized smoothing} has proven to be an effective countermeasure against adversarial attacks~\cite{cohen2019certified}. This technique (introduced in details in Section~\ref{subsection:randomized-smoothing}), perturbs the input data with random noise and returns an averaged prediction, thus attenuating the effect of any specific adversarial modification. The approach has been successfully applied in computer vision~\cite{salman2019provably,levine2020wasserstein} and language models~\cite{olivier2021sequential,seferis2025randomized}. However, to the best of our knowledge, it has not yet been applied to trajectory prediction.

In this work, we propose a methodology for applying randomized smoothing to arbitrary trajectory prediction models (as illustrated in Figure~\ref{fig:overview}) and assess its efficacy against realistic adversarial attacks. After introducing the concept of adversarial attacks on trajectory prediction and the \emph{randomized smoothing} technique (Sections~\ref{subsection:adv-attacks-trajectory-prediction}~and~\ref{subsection:randomized-smoothing}, respectively), we present two strategies for smoothing trajectory predictors in Section~\ref{sec:Smoothing_Traj}. Then, we empirically evaluate our proposed methods in Section~\ref{section:results}.
In particular, we generate adversarial attacks on two datasets (\emph{L-GAP}~\cite{zgonnikov_should_2022} and \emph{rounD}~\cite{krajewski_round_2020}) following~\cite{schumann2025realistic}, and evaluate the impact of the randomized smoothing (implemented following~\cite{rekavandi2024certified}) on established trajectory prediction models (\emph{Trajectron++}~\cite{salzmann_trajectron_2020} and \emph{ADAPT}~\cite{aydemir2023adapt}) using select robustness metrics.

\section{Background}

\subsection{Adversarial attacks on trajectory prediction models}\label{subsection:adv-attacks-trajectory-prediction}
In trajectory prediction, the goal is to forecast the future movement $Y_i = \left\{\bm{s}_{i}(t) \mid t \in \{\Delta t, \ldots, T \Delta t\} \right\}$, where $\bm{s} \in \mathbb{R}^D$ is recorded in intervals of $\Delta t$, of an agent $i$ over prediction horizon $T$ based on the past observed trajectories with observation horizon $H$ of all $N$ agents $\bm{X} = \{X_i \mid i \in \{1, \hdots, N\}\}$, where $X_i = \left\{\bm{s}_{i}(t) \mid t \in \{(-H+1) \Delta t, \ldots, 0\} \right\}$ are the trajectories of different agents and $\bm{s}$ is one state in a trajectory, including for example position or velocities. To this end, a probabilistic model $\mathds{P}_{\bm{\theta}}$ parametrized by $\bm{\theta} \in \bm{\Theta} \subseteq \mathbb{R}^Q$ is trained to generate $K$ trajectory samples $\widehat{Y}^k_i \sim \mathds{P}_{\bm{\theta}}(\cdot\vert X_i,\bm{X}_{\neg i})$ with $\widehat{\bm{Y}}_i=\{\widehat{Y}_i^k \mid k \in \{1, \ldots, K\} \}$ and $\bm{X}_{\neg i} = \{X_j \mid j \in \{1, \hdots, N\}\setminus\{i\}\}$.
Generally, such models are trained to minimize the distance between $\widehat{\bm{Y}}_i$ and ground truth observation $Y_i$, or to maximize the likelihood $\mathds{P}_{\bm{\theta}}(Y_i\vert X_i,\bm{X}_{\neg i})$~\cite{schumann2025realistic}\footnote{There are models predicting all agents simultaneously (e.g., see Rowe \emph{et al.}~\cite{rowe2023fjmp}), but we are not considering them here.}. 

An attack by an adversarial target agent ``tar'' (with index $i=\text{tar}$) then attempts to deceive the prediction model employed by the autonomous vehicle into making incorrect predictions. Specifically, the observed (past) states of the target agent are modified using perturbations $\delta_{X}$, resulting in the perturbed past trajectory $\widetilde{X}_{\text{tar}} = X_{\text{tar}} + \delta_{X}$ and the corresponding predictions $\widehat{\widetilde{Y}}_{\text{tar}} \sim \mathds{P}_\theta(\cdot|\widetilde{X}_{\text{tar}}, \bm{X}_{\neg \text{tar}})$, which can then used to evaluate the robustness of the models. 

While there have been many works on adversarial attacks on trajectory prediction models~\cite{cao_advdo_2022, cao_robust_2023, zhang2022adversarial, fan2024adversarial} suggesting different methods for generating $\delta_{X}$, in this work we will use the kinematically constrained attacks proposed by Schumann \emph{et al.}~\cite{schumann2025realistic}. This method was chosen because it is the only available approach that produces realistic attacks while preserving the original behavior of the target vehicle. In this ``white-box'' attack, projected gradient descent is used to generate an attack starting from the unperturbed input (i.e., $\delta_X^0 = \bm{0}$) in $M$ iterations. Abstractly, in each iteration $m \in \{1, \dots, M\}$ a gradient step (using step size $\alpha^m$) based on a loss function $\mathcal{L}$ is added to the previous displacement $\delta_X^{m-1}$ before a projection onto the feasible set $C$ (e.g., a hypercube around the unperturbed input $X$) is performed to match constraints:
\begin{equation}\begin{aligned}
    \widehat{\delta}_X^{m-1} & = \; \delta_X^{m-1} - \alpha^{m-1} \nabla_{\delta_X} \mathcal{L} \\
    \delta_X^{m} & = \; \underset{v \in C}{\mathrm{argmin}} \left\| v - \widehat{\delta}_X^{m-1} \right\|.
\end{aligned}\label{eq:PGD}
\end{equation}
A more detailed overview on the specific loss function $\mathcal{L}$ and the corresponding feasible set being used as a constraint $C$ can be found in the original work~\cite{schumann2025realistic}.

\subsection{Randomized smoothing}\label{subsection:randomized-smoothing}

\newtheorem{theorem}{Theorem}[section] 
\newtheorem{proposition}[theorem]{Proposition}
\newtheorem{example}{Example}
\newtheorem{lemma}[theorem]{Lemma}
\newtheorem{definition}{Definition}
\newtheorem{assumption}{Assumption}
\newtheorem{remark}{Remark}
\newtheorem{problem}{Problem}
\newtheorem{subproblem}{Subproblem}[problem]

\newcommand{\EF}[1]{\textcolor{red}{[EF: #1]}}

\def\realNum{{\mathbb{R}}}
\def\sX{{\mathcal{X}}}
\def\sY{{\mathcal{Y}}}
\def\sR{{\mathcal{R}}}
\def\vx{{{x}}}
\def\vy{{{y}}}
\def\Prob{{\mathbb{P}}}
\def\expect{{\mathbb{E}}}
\def\sN{\mathcal{N}}

\newcommand{\norm}[1]{\left\lVert#1\right\rVert}
\newcommand{\evy}[1]{{\vy^{(#1)}}}

\emph{Randomized Smoothing}~\cite{liu2018towards,lecuyer2019certified,cohen2019certified} has recently emerged as one of the primary techniques for improving the robustness of machine learning models. While the literature has initially focused on classification problems~\cite{cohen2019certified,li2019certified,yang2020randomized,levine2020wasserstein}, analogous results have been then presented for the regression setting~\cite{chiang2020detection,rekavandi2024rs,rekavandi2024certified}, which we briefly introduce in this Section. Let $f:\realNum^d \to \realNum^q$ be a \emph{base} regression model. The \emph{randomized smoothed} regression model is defined as
\begin{align}\label{def:randomized-smoothed-regressor}
    g(x)=\expect_\varepsilon \big[ f(x+\varepsilon) \big],
\end{align}
where $\varepsilon \sim \sN(\bm{0}, \sigma^2 I)$ is (typically) a Gaussian random noise with variance $\sigma^2 > 0$. Intuitively, Equation~\eqref{def:randomized-smoothed-regressor} represents a regressor averaging the value of the base regressor $f$ in a neighborhood around the input $x \in \realNum^d$. In practice, computing Equation~\eqref{def:randomized-smoothed-regressor} in closed-form is often infeasible~\cite{girard2002gaussian} (e.g. due to the potential non-linearity of $f$, which is the case, for instance, if $f$ is a neural network). Thus, this expression is generally replaced by the natural approximation
\begin{align}\label{eq:randomized_smoothin_approx}
    g_N(x) = \frac{1}{N} \sum_{i=1}^{N} f(x + \varepsilon_i), 
\end{align}
where $\{ \varepsilon_1,\dots,\varepsilon_N \}$ are i.i.d. samples from the distribution of $\varepsilon$. The main advantage of smoothed regressors is that they inherit some particular robustness properties against attacks in $L_p$-norm, without any need to retrain the base regressor \cite{rekavandi2024rs,rekavandi2024certified}. Indeed, Theorem 3 in~\cite{rekavandi2024certified} shows that, for a given input $x \in \realNum^d$, one may define a compact set $\sR$ that $g_N(x+\delta)$ belongs to with high probability for any sufficiently small attack $\delta$. In other words, by averaging predictions in a neighborhood of the point $x$, the smoothed regressor becomes less sensitive to inputs associated with high-norm gradients, thus regularizing the prediction output. The insight that averaging mechanisms can protect against adversarial attacks has also been observed in the context of Bayesian models \cite{bortolussi2024robustness}.

\section{Randomized smoothing for trajectory prediction}\label{sec:Smoothing_Traj}
After giving a short overview over both adversarial attacks against trajectory prediction models and randomized smoothing, we apply the latter to trajectory prediction.
Specifically, we first assume that for each agent $i$, the sampling of $K$ independent trajectories from the prediction distribution $\mathds{P}_{\bm{\theta}}(. \mid X_i, \bm{X}_{\neg i})$ described in Section \ref{subsection:adv-attacks-trajectory-prediction} induces a base (stochastic) prediction function\footnote{The dependence of $f_{\mathds{P}_{\bm{\theta}}}$ on the index $i$ and the other agents' trajectories $\bm{X}_{\neg i}$ is implicit for notational simplicity.} $f_{\mathds{P}_{\bm{\theta}}}: \mathbb{R}^{H\times D} \rightarrow \mathbb{R}^{K \times T \times D}$ such that
\begin{equation}\label{eq:pred_model}
    \widehat{\bm{Y}}_i = f_{\mathds{P}_{\bm{\theta}}}(X_i) \,.
\end{equation}
Furthermore, we assume that a trajectory state $\bm{s} = \{\bm{p},\bm{v},\bm{u}\} \in \mathbb{R}^D$ can be partitioned into a position $\bm{p}$, some auxiliary states $\bm{v}$, and the control inputs $\bm{u}$, where the latter are distinguished by being associated with the highest order of derivatives of $\bm{p}$ over time. Based on this, we then propose two approaches of randomized smoothing for the base prediction model $f_{\mathds{P}_{\bm{\theta}}}$ presented in Equation~\eqref{eq:pred_model}.

\subsection{Position-based smoothing}
Our first approach consists in smoothing the position coordinates of the vehicles' states, thus obtaining the smoothed predictor $g_N$ given by:
\begin{equation}\label{eq:smooth_pos}
    g_N(X) = \frac{1}{N}\sum\limits_{i=1}^{N} f_{\mathds{P}_{\bm{\theta}}}(X + \varepsilon_X)\,,\,\, \varepsilon_X\sim \mathcal{N}(0,\bm{\Sigma}_X),
\end{equation}
where $\bm{\Sigma}_X$ is only non-zero in the block-diagonal associated with the positions $\bm{p}$, to which we assign the variance $\sigma^2 > 0$.

\subsection{Control-based smoothing} 
Alternatively, under the assumption that there exists a dynamical model $\varphi$ such that $\{\bm{p}(t),\bm{v}(t)\} = \varphi(\bm{s}(t-\Delta t))$, we define the function $\Phi$ with $X_{\text{tar}} = \Phi(U_{\text{tar}})$, where $U_{\text{tar}}$
\begin{equation}\begin{aligned}
    U_{\text{tar}} = &\,\left\{\bm{p}_{\text{tar}}\left((-H+1)\Delta t\right), \bm{v}_{\text{tar}}\left((-H+1)\Delta t\right)\right\} \,\cup \\
    & \, \left\{\bm{u}_{\text{tar}}(t) \mid t \in \{(-H+1)\Delta t, \hdots, 0\} \right\}\,.
\end{aligned}
\end{equation}
We then propose a second smoothing approach, where the noise is applied to the control inputs, similarly to how the perturbations are applied to them in~\cite{schumann2025realistic}:
\begin{equation}\label{eq:smooth_ctrl}
    g_N(X) = \frac{1}{N}\sum\limits_{i=1}^{N} f_{\mathds{P}_{\bm{\theta}}}(\Phi(U + \varepsilon_U))\,,\,\, \varepsilon_U\sim \mathcal{N}(0,\bm{\Sigma}_U)
\end{equation}
This approach -- which could be described as a (higher-order) Ornstein–Uhlenbeck process~\cite{oksendal2013stochastic} -- is theorized to have the advantage that the resulting trajectories are less likely to display dynamically infeasible inputs (such as impossible sharp turns), and are therefore less prone to degrade model performance, if the model would rely on such properties.
The covariance $\bm{\Sigma}_U$ is only non-zero on the diagonal entries associated with the control inputs $\bm{u}$. Here, to allow for a fair comparison with the \emph{position-based} approach, the variance levels are chosen so that the noise accumulated after applying the dynamical model roughly matches the one expected from applying $\sigma^2$ onto the positions themselves. For example, for the acceleration, we would assume that the corresponding standard deviation expressed in $\bm{\Sigma}_U$ would be $\sigma_a = \sigma \frac{\sqrt{3}}{\sqrt{H^3}\Delta t^2}$.

\section{Evaluation}\label{section:results}
\subsection{Experiment setup}
To empirically evaluate the potential of randomized smoothing in the context of adversarial attacks against trajectory prediction models, we test the approaches introduced in Section \ref{sec:Smoothing_Traj} (position-based in Equation~\eqref{eq:smooth_pos} -- henceforth referred as \emph{pos} -- and control-based in Equation~\eqref{eq:smooth_ctrl} -- \emph{ctrl}) on two datasets and two base prediction models. For the datasets, following~\cite{schumann2025realistic}, we consider \emph{L-GAP}~\cite{zgonnikov_should_2022} and \emph{rounD}~\cite{krajewski_round_2020}. In \emph{L-GAP}~\cite{zgonnikov_should_2022} (a driving simulator dataset), a vehicle performing a left turn at an intersection is assumed to be adversarial (similar to Figure~\ref{fig:overview}); it performs a turn across the path of an oncoming vehicle (the ego agent). In \emph{rounD}~\cite{krajewski_round_2020} (a real-world dataset collected on German roundabouts), the ego agent is on the roundabout and the adversarial agent tries to enter the roundabout in front of the ego agent. For the base trajectory prediction models, we employ \emph{Trajectron++}~\cite{salzmann_trajectron_2020} and \emph{ADAPT}~\cite{aydemir2023adapt}, two state-of-the-art models which are trained on each of the datasets. To compensate for \emph{L-GAP}'s small size, the models evaluated on that dataset are trained on a combination of \emph{L-GAP} and \emph{NuScenes}~\cite{caesar_nuscenes_2020}.

The loss function $\mathcal{L}$ used for generating adversarial attacks (Equation~\ref{eq:PGD}) is defined as the average displacement error (ADE) of the prediction model. Perturbations are applied not to the trajectories themselves but to the underlying control inputs, subject to further constraints that ensure that the attack remains meaningful (see Schumann \emph{et al.}~\cite{schumann2025realistic} for a more detailed description). This approach was selected because it produces adversarial attacks that are kinematically feasible and not easily detectable by simple filters.

In each experiment, we run multiple attacks by varying the perturbation limit $d_{\max} \in \{\SI{0.25}{m}, \SI{0.5}{m}, \SI{1}{m}\}$. The ADE metric is then also used to assess the performance of randomized smoothing, which is performed with various levels of noise standard deviation, i.e. $\sigma \in \{ \SI{0.25}{m}, \SI{0.5}{m}, \SI{1}{m} \}$.

Further, we empirically evaluate the impact of the training strategy, i.e. either employing smoothing during the training phase of the base predictor\footnote{Specifically, we directly train the smoothed regressor using $N=1$, resampling the smoothing noise at each epoch, which maintains the computational effort similar to training the base predictor.} -- \emph{train $\&$ eval}, or the standard practice of smoothing only during evaluation -- \emph{eval}. In all experiments, evaluation is performed with $N=20$.

\subsection{Results and discussion}
\begin{table}
    \centering
    \caption{Average displacement error (in meters $m$) on \emph{L-GAP} for \emph{Trajectron++}. The bold values shows the best result in each column.}
    \vspace{-0.5cm}
    \begin{tabularx}{\linewidth}{|X|X|X|YYYY|}
\hline
 \multicolumn{3}{|c|}{Smoothing approach} & \multicolumn{4}{c|}{$d_{\max}$ [$m$]} \\ 
 \multicolumn{1}{|c}{Applied} & \multicolumn{1}{c}{Goal} & $\sigma$ [$m$] &  (0.0) & 0.25 & 0.5 & 1.0 \\ 
\hline 
\multicolumn{2}{|c|}{No Smoothing} & (0.0) & 0.473 & 0.663 & 0.844 & 2.057  \\ 
\hline
\multirow{6}{*}{eval} & \multirow{3}{*}{pos~\eqref{eq:smooth_pos}} & 0.25 & \textbf{0.471} & \textbf{0.637} & 0.833 & \textbf{2.020} \\ 
 &  & 0.5 & 0.546 & 0.761 & 0.902 & 2.063  \\ 
 &  & 1.0 & 0.676 & 0.850 & 1.015 & 2.123  \\ 
\cline{2-7}
 & \multirow{3}{*}{ctrl~\eqref{eq:smooth_ctrl}} & 0.25 & 0.637 & 0.749 & \textbf{0.762} & 2.563  \\ 
 &  & 0.5 & 0.862 & 0.888 & 0.961 & 2.410  \\ 
 &  & 1.0 & 1.237 & 1.232 & 1.323 & 2.331  \\ 
\hline
\end{tabularx}

    \vspace{-0.5cm}
    \label{tab:LGAP_Tpp}
\end{table}

\begin{table}
    \centering
    \caption{Average displacement error (in meters $m$) on \emph{L-GAP} for \emph{ADAPT}. The bold values shows the best result in each column.}
    \vspace{-0.5cm}
    \begin{tabularx}{\linewidth}{|X|X|X|YYYY|}
\hline
 \multicolumn{3}{|c|}{Smoothing approach} & \multicolumn{4}{c|}{$d_{\max}$ [$m$]} \\ 
 \multicolumn{1}{|c}{Applied} & \multicolumn{1}{c}{Goal} & $\sigma$ [$m$] &  (0.0) & 0.25 & 0.5 & 1.0 \\ 
\hline 
\multicolumn{2}{|c|}{No Smoothing} & (0.0) & 1.169 & 1.867 & 2.262 & 2.952  \\ 
\hline
\multirow{6}{*}{eval} & \multirow{3}{*}{pos~\eqref{eq:smooth_pos}} & 0.25 & 1.420 & 1.617 & 1.958 & 2.723  \\ 
 &  & 0.5 & 1.937 & 1.901 & 2.005 & 2.532  \\ 
 &  & 1.0 & 6.576 & 6.210 & 5.996 & 5.716  \\ 
\cline{2-7}
 & \multirow{3}{*}{ctrl~\eqref{eq:smooth_ctrl}} & 0.25 & \textbf{1.164} & 1.509 & \textbf{1.558} & \textbf{1.428} \\ 
 &  & 0.5 & 1.214 & \textbf{1.494} & 1.574 & 1.504  \\ 
 &  & 1.0 & 1.394 & 1.585 & 1.708 & 1.728  \\ 
\hline
\end{tabularx}

    \vspace{-0.5cm}
    \label{tab:LGAP_ADAPT}
\end{table}

The results in Tables~\ref{tab:LGAP_Tpp}, \ref{tab:LGAP_ADAPT}, \ref{tab:RounD_Tpp},~and~\ref{tab:RounD_ADAPT} suggest that, compared with the base prediction models, employing randomized smoothing improves models' ADE performance in all adversarial attack experiments (i.e., $d_{\max}>0$). Additionally, with the exception of \emph{Trajectron++} on \emph{rounD} (Table~\ref{tab:RounD_Tpp}), randomized smoothing (either position-based or control-based) achieved comparable performance on the unperturbed datasets as well (i.e, $d_{\max}=0$), and in one case (\emph{ADAPT} on \emph{rounD}; Table~\ref{tab:RounD_ADAPT}) even improved baseline performance. While earlier work in other domains demonstrated that randomized smoothing typically deteriorates performance in nominal scenarios~\cite{cohen2019certified, lecuyer2019certified}, our results support the idea that randomized smoothing can function as a form of regularization that protects against model overfitting~\cite{bishop_noise_1995}. 

Furthermore, we observe that randomized smoothing is capable of lowering the correlation between the perturbation limit $d_{\max}$ and the resulting ADE. For instance, when applying the control-based approach (\emph{ctrl}, Equation~\eqref{eq:smooth_ctrl}) with noise variance level $\sigma = \SI{0.25}{m}$ to the base model \emph{ADAPT}, the ADE differences between $d_{\max} = \SI{0.25}{m}$ and $d_{\max} = \SI{1}{m}$ are much less severe ($\SI{-0.81}{m}$ vs. $\SI{1.095}{m}$ on \emph{L-GAP} in Table~\ref{tab:LGAP_ADAPT} and $\SI{0.028}{m}$ vs. $\SI{0.92}{m}$ on \emph{rounD} in Table~\ref{tab:RounD_ADAPT}), which suggests a reduction of the impact of the attack magnitude on the model predictions in general.

\begin{table}
    \centering
    \caption{Average displacement error (in meters $m$) on \emph{rounD} for \emph{Trajectron++}. The bold values shows the best result in each column.}
    \vspace{-0.5cm}
    \begin{tabularx}{\linewidth}{|X|X|X|YYYY|}
\hline
 \multicolumn{3}{|c|}{Smoothing approach} & \multicolumn{4}{c|}{$d_{\max}$ [$m$]} \\ 
 \multicolumn{1}{|c}{Applied} & \multicolumn{1}{c}{Goal} & $\sigma$ [$m$] &  (0.0) & 0.25 & 0.5 & 1.0 \\ 
\hline 
\multicolumn{2}{|c|}{No Smoothing} & (0.0) & \textbf{0.116} & 4.989 & 5.562 & 6.335  \\ 
\hline
\multirow{6}{*}{eval} & \multirow{3}{*}{pos~\eqref{eq:smooth_pos}} & 0.25 & 0.341 & 1.405 & 1.610 & 2.708  \\ 
 &  & 0.5 & 0.520 & 1.430 & 1.697 & 2.708  \\ 
 &  & 1.0 & 0.614 & 1.467 & 1.872 & 2.766  \\ 
\cline{2-7}
 & \multirow{3}{*}{ctrl~\eqref{eq:smooth_ctrl}} & 0.25 & 0.533 & 2.822 & 3.025 & 4.880  \\ 
 &  & 0.5 & 0.607 & 5.422 & 5.663 & 6.194  \\ 
 &  & 1.0 & 0.793 & 10.604 & 10.817 & 10.088  \\ 
\hline
\multirow{6}{*}{\makecell[l]{train \& \\eval}} & \multirow{3}{*}{pos~\eqref{eq:smooth_pos}} & 0.25 & 0.300 & 1.171 & 1.251 & 1.724  \\ 
 &  & 0.5 & 0.477 & 1.357 & 1.481 & 2.273  \\ 
 &  & 1.0 & 0.817 & 1.412 & 1.510 & 2.031  \\ 
\cline{2-7}
 & \multirow{3}{*}{ctrl~\eqref{eq:smooth_ctrl}} & 0.25 & 0.376 & \textbf{0.487} & \textbf{0.538} & 1.104  \\ 
 &  & 0.5 & 0.498 & 0.540 & 0.559 & \textbf{0.930} \\ 
 &  & 1.0 & 0.916 & 0.892 & 0.889 & 0.943  \\ 
\hline
\end{tabularx}

    \vspace{-0.5cm}
    \label{tab:RounD_Tpp}
\end{table}

The effect of the training strategy is, however, less clear. For instance, employing randomized smoothing during the training phase (\emph{train $\&$ eval}) improves the results for \emph{Trajectron++} for both the position- (\emph{pos}) and control-based (\emph{ctrl}) approaches (Table~\ref{tab:RounD_Tpp}), while for \emph{ADAPT}, it is only beneficial for the position-based. 
Similarly, the impact of the smoothing approaches in Equations \eqref{eq:smooth_pos} and \eqref{eq:smooth_ctrl} is model-dependent\footnote{While we tried to make the impact of the similar values of $\sigma$ across the two approaches comparable, the validity of such a comparison cannot be guaranteed.}. Concretely, \emph{ADAPT}'s best results are generally achieved when smoothing control inputs (Tables~\ref{tab:LGAP_ADAPT}~and~\ref{tab:RounD_ADAPT}). In contrast, for \emph{Trajectron++}, if smoothing is only used for evaluation, the position-based approach \emph{pos}, with very few exceptions, presents superior results. Given the lack of clear preferences, future work might also try to explore the benefit of combining both approaches.

\begin{table}
    \centering
    \caption{Average displacement error (in meters $m$) on \emph{rounD} for \emph{ADAPT}. The bold values shows the best result in each column.}
    \vspace{-0.5cm}
    \begin{tabularx}{\linewidth}{|X|X|X|YYYY|}
\hline
 \multicolumn{3}{|c|}{Smoothing approach} & \multicolumn{4}{c|}{$d_{\max}$ [$m$]} \\ 
 \multicolumn{1}{|c}{Applied} & \multicolumn{1}{c}{Goal} & $\sigma$ [$m$] &  (0.0) & 0.25 & 0.5 & 1.0 \\ 
\hline 
\multicolumn{2}{|c|}{No Smoothing} & (0.0) & 0.537 & 0.655 & 0.654 & 0.747  \\ 
\hline
\multirow{6}{*}{eval} & \multirow{3}{*}{pos~\eqref{eq:smooth_pos}} & 0.25 & 1.504 & 1.582 & 1.563 & 1.644  \\ 
 &  & 0.5 & 2.308 & 2.343 & 2.347 & 2.383  \\ 
 &  & 1.0 & 3.541 & 3.554 & 3.563 & 3.565  \\ 
\cline{2-7}
 & \multirow{3}{*}{ctrl~\eqref{eq:smooth_ctrl}} & 0.25 & \textbf{0.494} & \textbf{0.541} & \textbf{0.552} & \textbf{0.569} \\ 
 &  & 0.5 & 0.567 & 0.612 & 0.621 & 0.636  \\ 
 &  & 1.0 & 0.789 & 0.827 & 0.835 & 0.847  \\ 
\hline
\multirow{6}{*}{\makecell[l]{train \& \\eval}} & \multirow{3}{*}{pos~\eqref{eq:smooth_pos}} & 0.25 & 1.174 & 1.185 & 1.173 & 1.206  \\ 
 &  & 0.5 & 1.420 & 1.427 & 1.415 & 1.443  \\ 
 &  & 1.0 & 1.670 & 1.677 & 1.665 & 1.688  \\ 
\cline{2-7}
 & \multirow{3}{*}{ctrl~\eqref{eq:smooth_ctrl}} & 0.25 & 0.647 & 0.669 & 0.676 & 0.685  \\ 
 &  & 0.5 & 1.183 & 1.178 & 1.178 & 1.187  \\ 
 &  & 1.0 & 1.924 & 1.915 & 1.911 & 1.920  \\ 
\hline
\end{tabularx}

    \vspace{-0.5cm}
    \label{tab:RounD_ADAPT}
\end{table}

Meanwhile, the effect of noise variance $\sigma^2$ is much more pronounced, and the smallest ADEs are achieved most often with the smallest values of $\sigma$ (the noticeable exceptions are for $d_{\max}=\SI{0.25}{m}$ in Table~\ref{tab:LGAP_ADAPT} and $d_{\max}=\SI{1}{m}$ in Table~\ref{tab:RounD_Tpp}). This seems reasonable, as lower variances likely allow for a better balance between increasing the prediction robustness while maintaining the local accuracy of the base model.

Importantly, given that randomized smoothing is an approach that promises increased robustness without retraining the base model, it is remarkable that, for both \emph{Trajectron++} and \emph{ADAPT}, we can find at least one smoothing setting which consistently improves on the base model across all datasets for $d_{\max}>0$. In particular, for \emph{Trajectron++}, this is the case for the position-based (\emph{pos}) approach when $\sigma=\SI{0.25}{m}$, while for \emph{ADAPT}, we can point to the control-based (\emph{ctrl}) approach also with $\sigma=\SI{0.25}{m}$.

While our results suggest that randomized smoothing is a promising technique for enhancing the robustness of trajectory prediction models, future work should explore whether the probabilistic robustness guarantees in~\cite{rekavandi2024certified} generate tight useful prediction sets in the case of trajectory prediction, while also further exploring their $\alpha$-smoothing technique. Furthermore, it is interesting to explore whether the variance of the randomized predictor (i.e., $f(x + \varepsilon)$) could be used as a reliable proxy for the uncertainty of the base predictor, which would be useful information for the motion planner using such predictions.
Finally, expanding to more models, datasets, and adversarial attack methods would improve the support for the claims in this section.

\section{Conclusion}
In this work, we explored randomized smoothing as a defense mechanism for trajectory prediction models against adversarial attacks.
Our results indicate that this technique consistently improves performance (i.e., reduced average displacement error) of trajectory predictions under adversarial attacks, while also lowering the model sensitivity to the magnitude of the adversarial perturbation. Interestingly, we observe that the smoothed predictors exhibit at most a marginal reduction in accuracy on the original, non-adversarial data, even achieving improved performance in some settings.
Consequently, randomized smoothing seems to be capable of consistently improving empirical robustness for the base models across datasets and perturbation magnitudes studied in this work. Even though our results indicate that the ideal smoothing approach is model-dependent, our work positions randomized smoothing as a computationally inexpensive and practical tool for increasing the robustness of trajectory prediction systems, and motivate future work on certified robustness guarantees in this domain.

\bibliographystyle{jabbrv_ieeetr}
\bibliography{manual}

@inproceedings{hagenus2024survey,
  title={Robustness in trajectory prediction for autonomous vehicles: a survey},
  author={Hagenus, Jeroen and Mathiesen, Frederik Baymler and Schumann, Julian F and Zgonnikov, Arkady},
  booktitle={2024 IEEE Intelligent Vehicles Symposium (IV)},
  pages={969--976},
  year={2024},
  organization={IEEE}
}

@article{tocchetti2022ai,
  title={Ai robustness: a human-centered perspective on technological challenges and opportunities},
  author={Tocchetti, Andrea and Corti, Lorenzo and Balayn, Agathe and Yurrita, Mireia and Lippmann, Philip and Brambilla, Marco and Yang, Jie},
  journal={ACM Computing Surveys},
  volume={57},
  number={6},
  pages={1--38},
  year={2025},
  publisher={ACM New York, NY}
}

@inproceedings{zhang2022adversarial,
  title={On adversarial robustness of trajectory prediction for autonomous vehicles},
  author={Zhang, Qingzhao and Hu, Shengtuo and Sun, Jiachen and Chen, Qi Alfred and Mao, Z Morley},
  booktitle={Proceedings of the IEEE/CVF Conference on Computer Vision and Pattern Recognition},
  pages={15159--15168},
  year={2022}
}

@inproceedings{nayakanti_wayformer_2023,
	title = {Wayformer: {Motion} {Forecasting} via {Simple} \& {Efficient} {Attention} {Networks}},
	shorttitle = {Wayformer},
	url = {https://ieeexplore.ieee.org/abstract/document/10160609},
	doi = {10.1109/ICRA48891.2023.10160609},
	urldate = {2024-12-02},
	booktitle = {2023 {IEEE} {International} {Conference} on {Robotics} and {Automation} ({ICRA})},
	author = {Nayakanti, Nigamaa and Al-Rfou, Rami and Zhou, Aurick and Goel, Kratarth and Refaat, Khaled S. and Sapp, Benjamin},
	month = may,
	year = {2023},
	pages = {2980--2987},
}

@article{montali_waymo_2023,
	title = {The {Waymo} {Open} {Sim} {Agents} {Challenge}},
	volume = {36},
	url = {https://proceedings.neurips.cc/paper_files/paper/2023/hash/b96ce67b2f2d45e4ab315e13a6b5b9c5-Abstract-Datasets_and_Benchmarks.html},
	language = {en},
	urldate = {2024-10-21},
	journal = {Advances in Neural Information Processing Systems},
	author = {Montali, Nico and Lambert, John and Mougin, Paul and Kuefler, Alex and Rhinehart, Nicholas and Li, Michelle and Gulino, Cole and Emrich, Tristan and Yang, Zoey and Whiteson, Shimon and White, Brandyn and Anguelov, Dragomir},
	month = dec,
	year = {2023},
	pages = {59151--59171},
}

@inproceedings{meszaros_trajflow_2024,
	address = {Jeju},
	title = {{TrajFlow}: {Learning} {Distributions} over {Trajectories} for {Human} {Behavior} {Prediction}},
	shorttitle = {{TrajFlow}},
	language = {en},
	booktitle = {2024 {IEEE} {Intelligent} {Vehicles} {Symposium} ({IV})},
	author = {Mészáros, Anna and Schumann, Julian F. and Alonso-Mora, Javier and Zgonnikov, Arkady and Kober, Jens},
	month = jun,
	year = {2024},
}

@article{schumann_benchmarking_2023,
	title = {Benchmarking {Behavior} {Prediction} {Models} in {Gap} {Acceptance} {Scenarios}},
	volume = {8},
	issn = {2379-8904},
	doi = {10.1109/TIV.2023.3244280},
	number = {3},
	journal = {IEEE Transactions on Intelligent Vehicles},
	author = {Schumann, Julian F. and Kober, Jens and Zgonnikov, Arkady},
	month = mar,
	year = {2023},
	pages = {2580--2591},
}

@article{zgonnikov_should_2022,
	title = {Should {I} {Stay} or {Should} {I} {Go}? {Cognitive} {Modeling} of {Left}-{Turn} {Gap} {Acceptance} {Decisions} in {Human} {Drivers}},
	issn = {0018-7208},
	shorttitle = {Should {I} {Stay} or {Should} {I} {Go}?},
	url = {https://doi.org/10.1177/00187208221144561},
	doi = {10.1177/00187208221144561},
	language = {en},
	urldate = {2023-02-08},
	journal = {Human Factors},
	author = {Zgonnikov, Arkady and Abbink, David and Markkula, Gustav},
	month = dec,
	year = {2022},
	note = {Publisher: SAGE Publications Inc},
	pages = {00187208221144561},
}

@article{mozaffari_deep_2022,
	title = {Deep {Learning}-{Based} {Vehicle} {Behavior} {Prediction} for {Autonomous} {Driving} {Applications}: {A} {Review}},
	volume = {23},
	issn = {1558-0016},
	shorttitle = {Deep {Learning}-{Based} {Vehicle} {Behavior} {Prediction} for {Autonomous} {Driving} {Applications}},
	doi = {10.1109/TITS.2020.3012034},
	number = {1},
	journal = {IEEE Transactions on Intelligent Transportation Systems},
	author = {Mozaffari, Sajjad and Al-Jarrah, Omar Y. and Dianati, Mehrdad and Jennings, Paul and Mouzakitis, Alexandros},
	month = jan,
	year = {2022},
	pages = {33--47},
}

@article{ho2020denoising,
  title={Denoising diffusion probabilistic models},
  author={Ho, Jonathan and Jain, Ajay and Abbeel, Pieter},
  journal={Advances in neural information processing systems},
  volume={33},
  pages={6840--6851},
  year={2020}
}

@book{oksendal2013stochastic,
  title={Stochastic differential equations: an introduction with applications},
  author={Oksendal, Bernt},
  year={2013},
  publisher={Springer Science \& Business Media}
}

@article{xu2025trajectory,
  title={Trajectory prediction meets large language models: A survey},
  author={Xu, Yi and Yang, Ruining and Zhang, Yitian and Lu, Jianglin and Zhang, Mingyuan and Wang, Yizhou and Su, Lili and Fu, Yun},
  journal={arXiv preprint arXiv:2506.03408},
  year={2025}
}

@inproceedings{seff2023motionlm,
  title={Motionlm: Multi-agent motion forecasting as language modeling},
  author={Seff, Ari and Cera, Brian and Chen, Dian and Ng, Mason and Zhou, Aurick and Nayakanti, Nigamaa and Refaat, Khaled S and Al-Rfou, Rami and Sapp, Benjamin},
  booktitle={Proceedings of the IEEE/CVF International Conference on Computer Vision},
  pages={8579--8590},
  year={2023}
}

@book{EU2019future,
author = {European Commission and Joint Research Centre},
title = {The future of road transport – Implications of automated, connected, low-carbon and shared mobility},
publisher = {Publications Office},
year = {2019},
doi = {doi/10.2760/668964}}

@article{duboz2025scenarios,
  title={Scenarios for the deployment of automated vehicles in Europe},
  author={Duboz, Louison and Raileanu, Ioan Cristinel and Krause, Jette and Norman-L{\'o}pez, Ana and Weitzel, Matthias and Ciuffo, Biagio},
  journal={Transportation Research Interdisciplinary Perspectives},
  volume={32},
  pages={101530},
  year={2025},
  publisher={Elsevier}
}

@inproceedings{seferis2025randomized,
  title={Randomized Smoothing Meets Vision-Language Models},
  author={Seferis, Emmanouil and Wu, Changshun and Kollias, Stefanos and Bensalem, Saddek and Cheng, Chih-Hong},
  booktitle={Proceedings of the 2025 Conference on Empirical Methods in Natural Language Processing},
  pages={27456--27466},
  year={2025}
}

@inproceedings{olivier2021sequential,
  title={Sequential Randomized Smoothing for Adversarially Robust Speech Recognition},
  author={Olivier, Raphael and Raj, Bhiksha},
  booktitle={Proceedings of the 2021 Conference on Empirical Methods in Natural Language Processing},
  pages={6372--6386},
  year={2021}
}

@article{salman2019provably,
  title={Provably robust deep learning via adversarially trained smoothed classifiers},
  author={Salman, Hadi and Li, Jerry and Razenshteyn, Ilya and Zhang, Pengchuan and Zhang, Huan and Bubeck, Sebastien and Yang, Greg},
  journal={Advances in neural information processing systems},
  volume={32},
  year={2019}
}

@article{schumann2025realistic,
title = {{Realistic} {Adversarial} {Attacks} for {Robustness} {Evaluation} of {Trajectory} {Prediction} {Models} via {Future} {State} {Perturbation}},
journal = {ACM Journal on Autonomous Transportation Systems},
author = {Schumann, Julian F. and Hagenus, Jeroen and Mathiesen, Frederik Baymler and Zgonnikov, Arkady},
year = {2026}
}

@inproceedings{bae2024singulartrajectory,
  title={Singulartrajectory: Universal trajectory predictor using diffusion model},
  author={Bae, Inhwan and Park, Young-Jae and Jeon, Hae-Gon},
  booktitle={Proceedings of the IEEE/CVF Conference on Computer Vision and Pattern Recognition},
  pages={17890--17901},
  year={2024}
}

@article{milford_self-driving_2020,
	title = {Self-{Driving} {Vehicles}: {Key} {Technical} {Challenges} and {Progress} {Off} the {Road}},
	volume = {39},
	issn = {1558-1772},
	shorttitle = {Self-{Driving} {Vehicles}},
	doi = {10.1109/MPOT.2019.2939376},
	number = {1},
	journal = {IEEE Potentials},
	author = {Milford, Michael and Anthony, Sam and Scheirer, Walter},
	month = jan,
	year = {2020},
	pages = {37--45},
}

@inproceedings{krajewski_round_2020,
	title = {The {rounD} {Dataset}: {A} {Drone} {Dataset} of {Road} {User} {Trajectories} at {Roundabouts} in {Germany}},
	shorttitle = {The {rounD} {Dataset}},
	doi = {10.1109/ITSC45102.2020.9294728},
	booktitle = {{IEEE} {International} {Conference} on {Intelligent} {Transportation} {Systems} ({ITSC})},
	author = {Krajewski, Robert and Moers, Tobias and Bock, Julian and Vater, Lennart and Eckstein, Lutz},
	month = sep,
	year = {2020},
	pages = {1--6},
}

@article{sinha_crash_2021,
	title = {Crash and disengagement data of autonomous vehicles on public roads in {California}},
	volume = {8},
	issn = {2052-4463},
	doi = {10.1038/s41597-021-01083-7},
	language = {en},
	number = {1},
	urldate = {2021-12-30},
	journal = {Scientific Data},
	author = {Sinha, Amolika and Chand, Sai and Vu, Vincent and Chen, Huang and Dixit, Vinayak},
	month = dec,
	year = {2021},
	pages = {298},
}

@misc{holland-letz_mobilitys_2021,
	title = {Mobility’s future: {An} investment reality check {\textbar} {McKinsey}},
	url = {https://www.mckinsey.com/industries/automotive-and-assembly/our-insights/mobilitys-future-an-investment-reality-check},
	urldate = {2022-05-02},
	author = {Holland-Letz, Daniel and Kässer, Matthias and Kloss, Benedikt and Müller, Thibaut},
	month = apr,
	year = {2021},
}

@inproceedings{aydemir2023adapt,
  title={Adapt: Efficient multi-agent trajectory prediction with adaptation},
  author={Aydemir, G{\"o}rkay and Akan, Adil Kaan and G{\"u}ney, Fatma},
  booktitle={Proceedings of the IEEE/CVF International Conference on Computer Vision},
  pages={8295--8305},
  year={2023}
}

@inproceedings{rowe2023fjmp,
  title={Fjmp: Factorized joint multi-agent motion prediction over learned directed acyclic interaction graphs},
  author={Rowe, Luke and Ethier, Martin and Dykhne, Eli-Henry and Czarnecki, Krzysztof},
  booktitle={Proceedings of the IEEE/CVF Conference on Computer Vision and Pattern Recognition},
  pages={13745--13755},
  year={2023}
}

@ARTICLE{bishop_noise_1995,
  author={Bishop, Chris M.},
  journal={Neural Computation}, 
  title={Training with Noise is Equivalent to Tikhonov Regularization}, 
  year={1995},
  volume={7},
  number={1},
  pages={108-116},
  doi={10.1162/neco.1995.7.1.108}
}

@inproceedings{fan2024novel,
  title={A Novel Unsupervised Anomaly Detection Method on Adversarial Attacks for Autonomous Vehicles Trajectory Prediction},
  author={Fan, Jiping and Wang, Zhenpo and Li, Guoqiang},
  booktitle={2024 IEEE 22nd International Conference on Industrial Informatics (INDIN)},
  pages={1--18},
  year={2024},
  organization={IEEE}
}

@inproceedings{fan2024adversarial,
  title={Adversarial Attack on Trajectory Prediction for Autonomous Vehicles with Generative Adversarial Networks},
  author={Fan, Jiping and Wang, Zhenpo and Li, Guoqiang},
  booktitle={2024 IEEE/RSJ International Conference on Intelligent Robots and Systems (IROS)},
  pages={1026--1031},
  year={2024},
  organization={IEEE}
}

@inproceedings{cao_robust_2023,
	title = {Robust {Trajectory} {Prediction} against {Adversarial} {Attacks}},
	url = {https://proceedings.mlr.press/v205/cao23a.html},
	language = {en},
	urldate = {2024-04-11},
	booktitle = {Proceedings of {The} 6th {Conference} on {Robot} {Learning}},
	publisher = {PMLR},
	author = {Cao, Yulong and Xu, Danfei and Weng, Xinshuo and Mao, Zhuoqing and Anandkumar, Anima and Xiao, Chaowei and Pavone, Marco},
	month = mar,
	year = {2023},
	note = {ISSN: 2640-3498},
	pages = {128--137},
}

@inproceedings{salzmann_trajectron_2020,
	address = {Cham},
	title = {Trajectron++: {Dynamically}-{Feasible} {Trajectory} {Forecasting} with {Heterogeneous} {Data}},
	isbn = {978-3-030-58523-5},
	shorttitle = {Trajectron++},
	doi = {10.1007/978-3-030-58523-5_40},
	language = {en},
	booktitle = {Computer {Vision} – {ECCV} 2020},
	publisher = {Springer International Publishing},
	author = {Salzmann, Tim and Ivanovic, Boris and Chakravarty, Punarjay and Pavone, Marco},
	editor = {Vedaldi, Andrea and Bischof, Horst and Brox, Thomas and Frahm, Jan-Michael},
	year = {2020},
	pages = {683--700},
}

@inproceedings{caesar_nuscenes_2020,
	title = {{nuScenes}: {A} {Multimodal} {Dataset} for {Autonomous} {Driving}},
	isbn = {978-1-72817-168-5},
	shorttitle = {{nuScenes}},
	url = {https://www.computer.org/csdl/proceedings-article/cvpr/2020/716800l1618/1m3nGHQO3HW},
	doi = {10.1109/CVPR42600.2020.01164},
	language = {English},
	urldate = {2024-04-11},
	publisher = {IEEE Computer Society},
	author = {Caesar, Holger and Bankiti, Varun and Lang, Alex H. and Vora, Sourabh and Liong, Venice Erin and Xu, Qiang and Krishnan, Anush and Pan, Yu and Baldan, Giancarlo and Beijbom, Oscar},
	month = jun,
	year = {2020},
	pages = {11618--11628},
}

@inproceedings{cao_advdo_2022,
	address = {Cham},
	title = {{AdvDO}: {Realistic} {Adversarial} {Attacks} for {Trajectory} {Prediction}},
	isbn = {978-3-031-20065-6},
	shorttitle = {{AdvDO}},
	doi = {10.1007/978-3-031-20065-6_3},
	language = {en},
	booktitle = {Computer {Vision} – {ECCV} 2022},
	publisher = {Springer Nature Switzerland},
	author = {Cao, Yulong and Xiao, Chaowei and Anandkumar, Anima and Xu, Danfei and Pavone, Marco},
	editor = {Avidan, Shai and Brostow, Gabriel and Cissé, Moustapha and Farinella, Giovanni Maria and Hassner, Tal},
	year = {2022},
	pages = {36--52},
}

@article{schumann2025step,
  title={STEP: Structured Training and Evaluation Platform for benchmarking trajectory prediction models},
  author={Schumann, Julian F and M{\'e}sz{\'a}ros, Anna and Kober, Jens and Zgonnikov, Arkady},
  journal={arXiv preprint arXiv:2509.14801},
  year={2025}
}

@inproceedings{liu2018towards,
  title={Towards robust neural networks via random self-ensemble},
  author={Liu, Xuanqing and Cheng, Minhao and Zhang, Huan and Hsieh, Cho-Jui},
  booktitle={Proceedings of the european conference on computer vision (ECCV)},
  pages={369--385},
  year={2018}
}

@inproceedings{lecuyer2019certified,
  title={Certified robustness to adversarial examples with differential privacy},
  author={Lecuyer, Mathias and Atlidakis, Vaggelis and Geambasu, Roxana and Hsu, Daniel and Jana, Suman},
  booktitle={2019 IEEE symposium on security and privacy (SP)},
  pages={656--672},
  year={2019},
  organization={IEEE}
}

@article{li2019certified,
  title={Certified adversarial robustness with additive noise},
  author={Li, Bai and Chen, Changyou and Wang, Wenlin and Carin, Lawrence},
  journal={Advances in neural information processing systems},
  volume={32},
  year={2019}
}

@inproceedings{cohen2019certified,
  title={Certified adversarial robustness via randomized smoothing},
  author={Cohen, Jeremy and Rosenfeld, Elan and Kolter, Zico},
  booktitle={international conference on machine learning},
  pages={1310--1320},
  year={2019},
  organization={PMLR}
}

@inproceedings{levine2020wasserstein,
  title={Wasserstein smoothing: Certified robustness against wasserstein adversarial attacks},
  author={Levine, Alexander and Feizi, Soheil},
  booktitle={International conference on artificial intelligence and statistics},
  pages={3938--3947},
  year={2020},
  organization={PMLR}
}

@inproceedings{yang2020randomized,
  title={Randomized smoothing of all shapes and sizes},
  author={Yang, Greg and Duan, Tony and Hu, J Edward and Salman, Hadi and Razenshteyn, Ilya and Li, Jerry},
  booktitle={International conference on machine learning},
  pages={10693--10705},
  year={2020},
  organization={PMLR}
}

@article{rekavandi2024rs,
  title={RS-Reg: Probabilistic and robust certified regression through randomized smoothing},
  author={Rekavandi, Aref Miri and Ohrimenko, Olga and Rubinstein, Benjamin IP},
  journal={arXiv preprint arXiv:2405.08892},
  year={2024}
}

@article{rekavandi2024certified,
  title={Certified Adversarial Robustness via Randomized $\alpha$-Smoothing for Regression Models},
  author={Rekavandi, Aref and Farokhi, Farhad and Ohrimenko, Olga and Rubinstein, Benjamin},
  journal={Advances in Neural Information Processing Systems},
  volume={37},
  pages={134127--134150},
  year={2024}
}

@article{chiang2020detection,
  title={Detection as regression: Certified object detection with median smoothing},
  author={Chiang, Ping-yeh and Curry, Michael and Abdelkader, Ahmed and Kumar, Aounon and Dickerson, John and Goldstein, Tom},
  journal={Advances in Neural Information Processing Systems},
  volume={33},
  pages={1275--1286},
  year={2020}
}

@article{girard2002gaussian,
  title={Gaussian process priors with uncertain inputs application to multiple-step ahead time series forecasting},
  author={Girard, Agathe and Rasmussen, Carl and Candela, Joaquin Q and Murray-Smith, Roderick},
  journal={Advances in neural information processing systems},
  volume={15},
  year={2002}
}

@article{bortolussi2024robustness,
  title={On the robustness of bayesian neural networks to adversarial attacks},
  author={Bortolussi, Luca and Carbone, Ginevra and Laurenti, Luca and Patane, Andrea and Sanguinetti, Guido and Wicker, Matthew},
  journal={IEEE Transactions on Neural Networks and Learning Systems},
  volume={36},
  number={4},
  pages={6679--6692},
  year={2024},
  publisher={IEEE}
}

\end{document}